\documentclass{article}

\usepackage{PRIMEarxiv}

\usepackage[utf8]{inputenc} 
\usepackage[T1]{fontenc}    
\usepackage{hyperref}       
\usepackage{url}            
\usepackage{booktabs}       
\usepackage{amsfonts}       
\usepackage{nicefrac}       
\usepackage{microtype}      
\usepackage{lipsum}
\usepackage{fancyhdr}       
\usepackage{graphicx}       
\graphicspath{{media/}}     

\newcommand{\ie}{\textit{i}.\textit{e}. }
\newcommand{\eg}{\textit{e}.\textit{g}. }
\usepackage{amsthm}
\usepackage{algorithm} 
\usepackage{algpseudocode} 
\usepackage{soul}
\usepackage{xcolor}

\title{Explaining Generative Diffusion Models via Visual Analysis for Interpretable Decision-Making Process
\thanks{\textit{\underline{DOI}}: 
\textbf{\url{https://www.sciencedirect.com/science/article/pii/S0957417424000964}}} 
}

\author{
  Ji-Hoon Park, Yeong-Joon Ju, Seong-Whan Lee \\
  Department of Artificial Intelligence \\
  Korea University, \\
  Anam-dong, Seongbuk-gu,
Seoul, 02841, Republic of Korea\\
  \texttt{jhoon\_park@korea.ac.kr, yj\_ju@korea.ac.kr, sw.lee@korea.ac.kr} \\
}

\begin{document}
\maketitle

\begin{abstract}
Diffusion models have demonstrated remarkable performance in generation tasks.
Nevertheless, explaining the diffusion process remains challenging due to it being a sequence of denoising noisy images that are difficult for experts to interpret.
To address this issue, we propose the three research questions to interpret the diffusion process from the perspective of the visual concepts generated by the model and the region where the model attends in each time step.
We devise tools for visualizing the diffusion process and answering the aforementioned research questions to render the diffusion process human-understandable.
We show how the output is progressively generated in the diffusion process by explaining the level of denoising and highlighting relationships to foundational visual concepts at each time step through the results of experiments with various visual analyses using the tools.
First, we rigorously examine spatial recovery levels to understand a model's focal region during denoising concerning semantic content and detailed levels.
In doing so, we illustrate that the denoising model initiates image recovery from the region containing semantic information and progresses toward the area with finer-grained details.
Secondly, we explore how specific concepts are highlighted at each denoising step by aligning generated images with the prompts used to produce them. 
By observing the internal flow of the diffusion process, we show how a model strategically predicts a particular visual concept at each denoising step to complete the final image.
Finally, we extend our analysis to decode the visual concepts embedded in all the time steps of the process.
Throughout the training of the diffusion model, the model learns diverse visual concepts corresponding to each time-step, enabling the model to predict varying levels of visual concepts at different stages.
We substantiate our tools using Area Under Cover (AUC) score, correlation quantification, and cross-attention mapping.
Our findings provide insights into the diffusion process and pave the way for further research into explainable diffusion mechanisms.
\end{abstract}


\section{Introduction}
Deep learning models have achieved remarkable progress, notably in domains like computer vision, natural language processing, and speech synthesis.
Nevertheless, when applied to critical domains such as finance~\cite{kim2023combining}, weather phenomena~\cite{he2024data}, time-series problems~\cite{jo2023neural}, artistic expression~\cite{sigaki2018history,perc2020beauty}
and medical applications~\cite{brito2023fault,miguel2023analysis}, errors made by the models lead to severe consequences.
Therefore, various approaches to scrutinizing and ascertaining the underlying causes through analysis and explanations are important for the expert systems.
Notably, the diffusion process generates high-fidelity images via multiple denoising steps for images initialized with Gaussian noise, alleviating existing drawbacks of generative models~\cite{ho2020denoising,song2020denoising,zhang2023hierarchical}.
Additionally, numerous studies have successfully addressed downstream problems by applying the effectiveness of denoising diffusion models~\cite{ruiz2023dreambooth,kawar2023imagic,avrahami2022blended,kingma2021variational,li2023controlled,li2023fusiondiff}.
The diffusion process for image generation is divided into two processes: a forward process and a reverse process.
In the forward process, the diffusion model (\ie U-Net backbone) learns to predict added noise from noisy images, which are derived by adding Gaussian noise to target images.
In the reverse process, the diffusion model repeatedly estimates noise to be eliminated to synthesize desired images.
Recently, existing works~\cite{choi2022perception,kwon2022diffusion} have improved the diffusion process by focusing on the time-step issues.
\cite{choi2022perception} revealed that each stage in the denoising steps has the level of visual concepts by leveraging the degree of Signal-to-Noise Ratio (SNR). Specifically, the early steps and the latter steps generate high-level content (semantic information of the image) and low-level content (details of the image), respectively.
They~\cite{choi2022perception,kwon2022diffusion} devise the training method for the diffusion model by weighting certain steps or layers containing semantic information.
Although the approaches show improved performances, the diffusion process still makes mistakes, such as unrealistic images, artifacts, biases, and drops for required concepts~\cite{srinivasan2021biases,luccioni2023stable}.
Therefore, the decision-making process of the diffusion model should be interpretable to trust the outcomes of the algorithm~\cite{ding2022explainability} or understand the causes of the occurrences, before mitigating the limitations.
However, it is challenging for humans to comprehend the generative manner due to the denoising process predicting noise from noisy images and a black-box attribute of the diffusion model.
In generating the ultimate image, the diffusion model systematically eliminates Gaussian noise through iterations.
Therefore, comprehending and explaining the decision-making steps of the model within the denoising process is crucial for exploring the roots of the aforementioned issues.
\begin{figure}
  \centering
  \includegraphics[width=\textwidth]{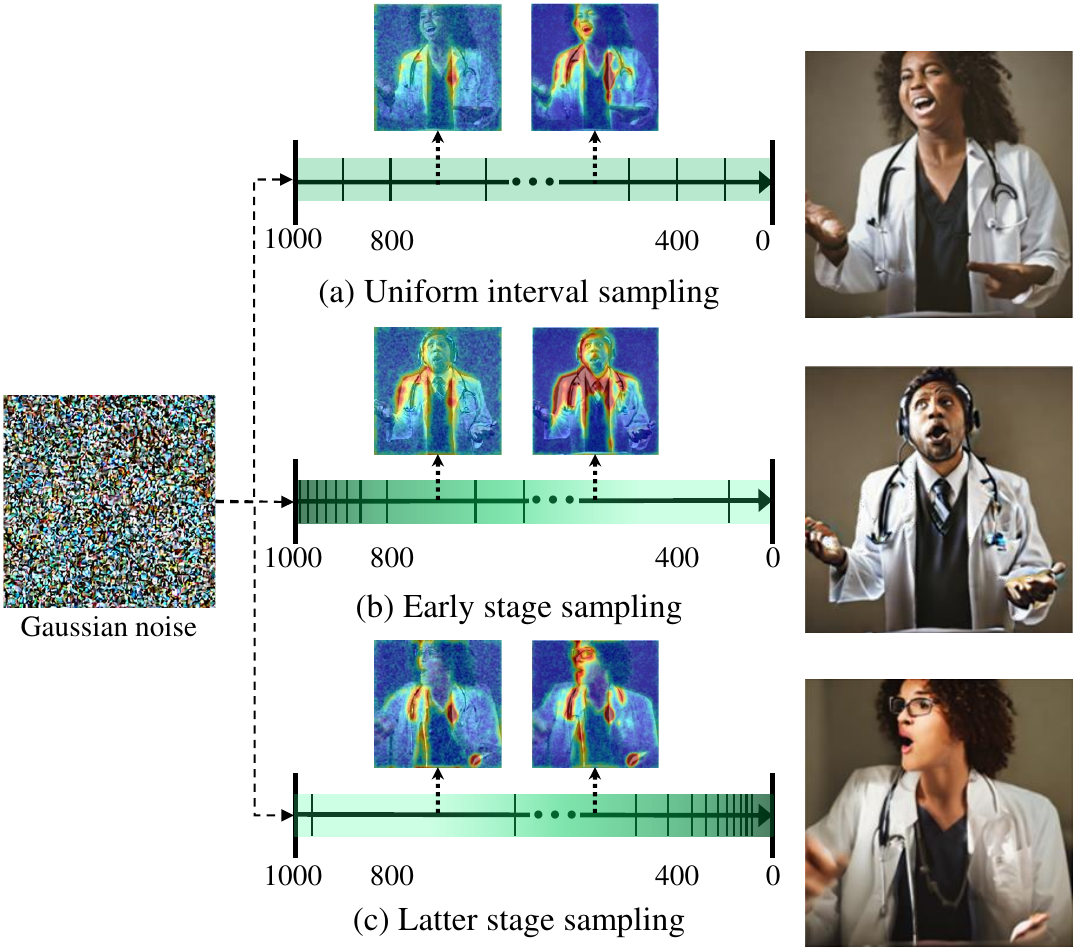}
  \caption{\textbf{Visualization of the image using saliency map and exponential sampling.} The images are generated from the prompt of ``a doctor singing a song". (a) Image generated using the original scheduler sampling which has the same interval. (b) image focusing on the early stage (\eg from $1000$ to $800$). (c) Image focusing on the latter stage (\eg from $200$ to $0$). The heat map shows the decision-making process of models in a particular step.}
  \label{fig:1}
\end{figure}
As illustrated in Fig.~\ref{fig:1}, it is difficult to be aware of when and how the concepts in a text condition are created during the process of generating images (\eg a headphone from ``singing" by a prompt ``a doctor singing a song".)
To explicitly investigate the diffusion process, we introduce three following questions concentrating on the effects observed at each time-step:

R1. \textbf{What regions are recovered by the diffusion model in terms of semantic and detail level?}
The denoising level indicates the region that the model recovers to generate the final image.
Specifically, the semantic level indicates a part of a concept that needs to be generated preferentially for image generation (\eg head of an object, structure of an object).
On the other hand, the detail level of the denoising process denotes refining fine-grained regions after constructing the structure of concepts by the semantic level of recovery.
The semantic and detail level of features in the denoising process is unidentifiable based only on the ratio of the noise.
To understand and interpret the diffusion process, users should observe the characteristics of image features in the images at each denoising step.
However, when we take an output image at a particular step, it is unidentifiable because of the noisy image.
Thus, we utilize the image's input and output to create a saliency map to display the region that the model concentrates on.
Thus, we visualize the region that the model focuses on by saliency map using the input and output of the image.
For visualizing the denoising process, we introduce the DiFfusion Randomized Input Sampling Explanation (DF-RISE), devised from RISE~\cite{petsiuk2018rise} since the RISE has limitations to employing generative models.
Thus, we implement a masking method and a similarity function for DF-RISE.
We analyze the level of the denoising process by visualizing all steps of the inference process with the heatmap.

R2. \textbf{Which specific concepts are prioritized at each time-step in order to generate images that align with the given conditional prompt?}
Although we interpret the denoising level in each step by visualizing the external of the model (\ie DF-RISE), the concepts from the conditional prompt to which the model attends remai incomprehensible.
Thus, to observe the internal workings of the model to understand the decision-making process of the diffusion model, we employ the DiFfusion gradient-weighted Class Activation Mapping (DF-CAM) devised from Grad-CAM~\cite{selvaraju2017grad} for applicability to the generative model.
We analyze the visual concepts from the prompt aligned to the diffusion model in each step using the internal visualization method, which visualizes the flow of gradient and activation maps.
The visualization tool shows features that the model concentrates on at during the denoising process in each step, through the heatmap of the noisy images.
Moreover, we qualify the activated concept by comparing the saliency map with the relevance score.

R3, \textbf{What visual concept is implied at time-step $t$?}
We extend our analysis beyond the concept level only in the inference step, to investigate the visual concepts embedded within each time stage across the entire sequence of time-steps $T$.
The time stage interpretation exhibits the visual concept the model concentrates on in each stage.
However, we cannot identify the difference in visual concept from the existing time-step sampling in inference, which has a uniform interval since the sampling was performed with the same distribution at each time stage as illustrated in Fig.~\ref{fig:1}(a).
Unlike such methods, we adjust time-steps sampling in the diffusion scheduler to control the interval of time-steps via the parallel transition of an exponential function named exponential sampling.
To compare attributes focused in each stage, exponential sampling typically divides the time-step of the diffusion process into several stages: the early stage (\ie a shorter interval from $1000$ to $800$ steps, such as Fig.~\ref{fig:1}(b)) and the latter stage (\ie a shorter interval from $200$ to $0$ steps, such as Fig.~\ref{fig:1}(c)), as illustrated in Fig.~\ref{fig:1}.
For instance, the attribute of ``gender" is generated in the early stage in Fig.~\ref{fig:1}(b) while the attribute is not determined in the latter stage in Fig.~\ref{fig:1}(c).
The attribute ``singing" in Fig.~\ref{fig:1}(b) (a man with headphones) is more emphasized than in the generated image through the scheduler with uniform intervals in Fig.~\ref{fig:1}(a).
However, the interpretation comparing each output can lead to various subjective interpretations among users.
Therefore, we implement the relevance score between the image and the prompt from the diffusion model to quantify the visual concept entailed in each step.
By addressing the research questions, we interpret the influence of time-step from the analysis that each step entails the information of the denoising level and the relation of the visual concept that is the rationale of the final output.

Moreover, we conducted various experiments to reveal the suitability and faithfulness of the tools.
First, we evaluate the performance of visualization methods with the Area Under the Curve (AUC) scores of deletion and insertion games.
Second, we quantify a correlation between the heat map region and the visual concept by comparing it with the attention score of the model.
Third, we prove the visual concept level using a cross-attention map composed of the relevance of image representation and conditional prompt.
Furthermore, through diverse experimental designs using various tools, we analyze the decisions made by the model for the final image and the visual concept of the time stage to enhance human understanding.
This analysis empowers users to comprehend the image generation process, moreover, our interpretations and tools will be instrumental in resolving issues such as model bias or artifacts.

\begin{itemize}
    \item We visualized the semantic and detail levels of the denoising process using DF-RISE. The diffusion model predicts the semantic elements in the early denoising step and then diffuses them to a wide range for the detail process.
    \item We interpreted the different visual concepts at each denoising step using DF-CAM. In each denoising step, the model predicts different visual concepts to generate a final high-fidelity image. The internal working visualization method visualizes the concepts that the model attends to.
    \item We analyzed the visual concepts entailed in time steps using exponential sampling and relevance scores. The different visual concepts in each time step $t$ can be explained by the relevance of the prompt and image and the intensively sampled output.
\end{itemize}

\label{introduction}

\section{Related Works}
\subsection{Denoising diffusion models}
The diffusion process outperforms various generative models, particularly in computer vision models. However, there are still areas that require improvement, such as high-resolution image generation and the time cost of inference. Various studies to address these problems can be classified into two types: conditional and unconditional. Denoising Diffusion Probabilistic Models (DDPM)~\cite{ho2020denoising} becomes the basis of the unconditioned diffusion process by formalizing diffusion models. \cite{song2020denoising} proposes Denoising Diffusion Implicit Models (DDIM) using a non-Markovian process to sample faster without quality degradation. Moreover, \cite{bond2022unleashing} reduces the number of denoising steps using vector quantization and transformer models for a two-stage process. In contrast, conditional diffusion model studies that apply the desired condition information are handled to solve various tasks. \cite{bordes2021high} trains by conditioning the representation from a different source, \cite{pandey2021vaes} constructs a generator-refiner framework by conditioning the Variational AutoEncoder (VAE), and \cite{ho2022cascaded} conditions the image classes for high-resolution image generation. These conditional diffusion models exhibit state-of-the-art performance, particularly in text-to-image~\cite{ruiz2023dreambooth,kawar2023imagic} and image editing~\cite{avrahami2022blended,kingma2021variational}. In this paper, we analyze the diffusion process from Explainable Artificial Intelligence (xAI) perspective. Specifically, deviating from the performance of the diffusion model, we interpret the generating process step-by-step to allow the model to be human-understandable.
The noise scheduler is used to noise the data distribution or sampling time-steps during the diffusion process. In the case of \cite{ho2020denoising}, they use a linear noise scheduler that linearly increases the noise scale, and \cite{nichol2021improved} proposes a cosine scheduler to control the noise of the images. In this paper, we propose time-steps sampling for the scheduler in the reverse denoising process. \cite{kingma2021variational} further specifies the scheduler that implements SNR. Moreover, for the perspective of investigation diffusion model, \cite{choi2022perception} analyzes each step of the diffusion process using SNR where each step has different visual concepts.
Furthermore, \cite{kwon2022diffusion} investigates semantic latent space named h-space as the eighth layer for h-space and implements the space to improve image editing. These studies focus on model investigations for outperforming generation tasks. However, the purpose of our research is to interpret the diffusion generation process as human-interpretable by providing a separate visual explanation of different visual concepts in the denoising process.

\subsection{Visual saliency explanation}
There are various proposals to explain black-box Deep Neural Networks with visualization without a training process~\cite{gunning2019xai,feng2023analytical,xu2023explainable,feng2023vs,ding2022visualizing}. We can distinguish between two groups of methods regarding the perspective of inside the model. First, \cite{bach2015pixel,montavon2017explaining} visualize the model by calculating the importance of each layer through model back-propagation. Grad-CAM~\cite{selvaraju2017grad} computes the saliency map with gradient and activation maps from the classification target and visualizes the attributes on which the layer concentrates. Second, apart from the inside of the black-box models,
RISE~\cite{petsiuk2018rise} explains the model by estimating important attributes for classification using similarity score and randomly generated input mask. However, these works only specialize in classifiers. D-RISE~\cite{petsiuk2021black} applies RISE to object detection using similarity functions such as Intersection over Union (IoU) and cosine similarity. In this paper, inspired by the works~\cite{ju2022complete,petsiuk2018rise,petsiuk2021black}, we propose a novel masking method and a score function specialized in feature maps for the visualization of the diffusion process.
\label{related_works}

\section{Methods} 
We aimed to interpret the decision-making process of the generative diffusion model. In Sec.~\ref{method1}, we introduce the tools of the saliency map to highlight regions contributing to the estimation of noise in a denoising step. 
In Sec.~\ref{method2}, we describe the exponential time-steps sampling for the scheduler in detail.
\subsection{Preliminary}
\noindent\textbf{Denoising diffusion models.} 
The forward process refers to converting high dimensional data distribution into Gaussian distribution and restoring it from noise \cite{ho2020denoising}. The process implements a gradual Gaussian transition with the Markov Chain over time-step:
\begin{equation}\label{eq1}
    q(\mathbf{x}_t|\mathbf{x}_{t-1})=\mathcal{N}(\mathbf{x}_t;\sqrt{1-\beta_t}\mathbf{x}_{t-1}, \beta_t \mathbb{I}), 
\end{equation}
where $\beta$ denotes the variance scheduler for noise scaling at time-step $t$.
The noise sampled from $\mathbf{x_0}$ can be expressed by the following equation:
\begin{equation}\label{eq2}
    \mathbf{x}_t=\sqrt{\alpha_t}\mathbf{x}_0 + \sqrt{1-\alpha_t}\epsilon, 
\end{equation}
where $\epsilon \sim \mathcal{N}(0, \mathbb{I})$ and $\alpha_t$ is the cumulative product of $1-\beta_t$.
The reverse process recovers the noisy image from Gaussian noise by repeatedly removing noise obtained from a model $\epsilon_t^{\theta}$ that is trained for predicting noise through respective steps. The process begins from the Gaussian noise $\mathbf{x}_t \sim \mathcal{N}(0, \mathbb{I})$ as follows:
The reverse diffusion process recovers the noisy image from Gaussian noise by repeatedly removing noise obtained from a model $\epsilon_t^{\theta}$ at sampled time-step $t$, which is trained to predict noise through the respective steps. The process begins with Gaussian noise $\mathbf{x}_t \sim \mathcal{N}(0, \mathbb{I})$ as follows:
\begin{equation}\label{eq3}
    \mathbf{x}_{t-1}=\frac{1}{\sqrt{1-\beta_t}}(\mathbf{x}_t-\frac{\beta_t}{\sqrt{1-\alpha_t}}\epsilon_t^{\theta}(\mathbf{x}_t))+\sigma_t\mathbf{z}_t, 
\end{equation}
where $\mathbf{z}_t \sim \mathcal{N}(0, \mathbb{I})$, $\alpha$ and $\sigma_t^2$ denote the variance scheduler and a variance, separately. $\beta$ denotes the variance scheduler for noise scaling at time-step $t$
On the other hand, DDIM~\cite{song2020denoising} utilizes a deterministic reverse process of the non-Markov process, which is derived from the existing Markov process as follows:
\begin{equation}\label{eq4}
        \mathbf{x}_{t-1} = \sqrt{\alpha_{t-1}}\underbrace{(\frac{\mathbf{x}_t - \sqrt{1-\alpha_t}\epsilon_t^{\theta}(\mathbf{x}_t)}{\sqrt{\alpha_t}})}_{\textnormal{``predicted $x_0$"}}
        +\underbrace{\sigma_t\mathbf{z}_t}_{\textnormal{random noise}}
        \\+\underbrace{\sqrt{1-\alpha_{t-1}-\sigma_t^2}\cdot \epsilon_t^{\theta}(x_t)}_{\textnormal{``direction pointing to $\mathbf{x}_t$"}}.
\end{equation}

\subsection{Saliency map visualization for the diffusion generation process}\label{method1}
The visualization of regions the model attended in each denoising step is necessary, given that the diffusion process generates images incrementally through multiple steps rather than all at once. 
Existing methods~\cite{choi2022perception,kwon2022diffusion} reveal that the early denoising steps characterized by higher SNR recover a high level of semantic feature, and the latter steps with a lower ratio restore a low level of denoising process.
Despite this understanding, users still face challenges in comprehending both the semantic or detail level of features recovered by the model and the features to which each layer of the U-Net model attends, since the flow of input and output of the process is noise that humans cannot identify.
First, what is the semantic and detail level of the denoising process? 
Second, what features are emphasized at each time-step to generate images that align with the conditions (\ie the prompt)?
In order to address the limitations and answer the research questions, we propose leveraging saliency visualization tools to gain insights into the decision-making process of the diffusion model.
Specifically, considering the scarcity of methods designed specifically for generative models, we introduce and compare two techniques probing the internals and externals of the model (\ie Grad-CAM and RISE), to devise a more comprehensive understanding of the generative model's behavior.

\begin{figure*}
\begin{center}
\includegraphics[width=1.0\linewidth]{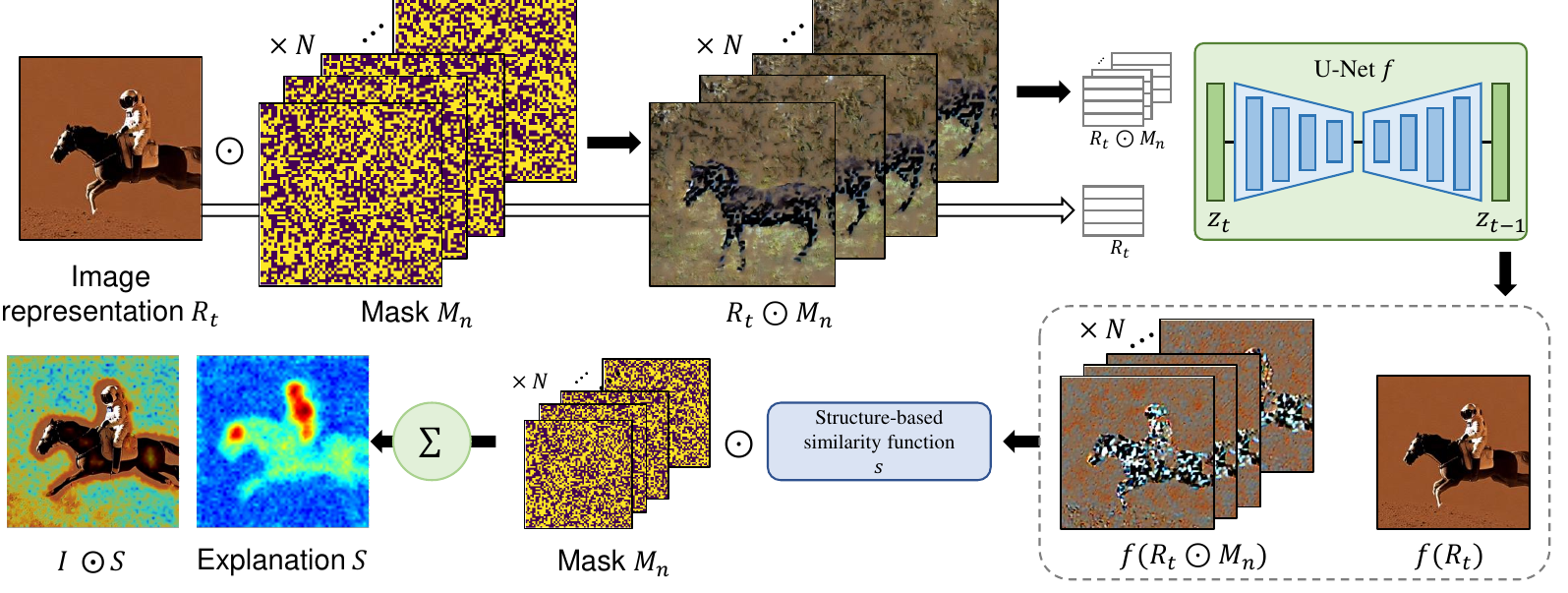}
\end{center}
\caption{\textbf{Overview of our proposed DF-RISE framework.}
 The framework includes a masking method and a structural similarity function that is applicable to the diffusion generative model. The saliency map is expressed using heatmaps. }
\label{fig:2}
\end{figure*}
\noindent\textbf{DF-RISE: Diffusion Randomized Input Sampling Explanation.}
The external of the model is visualized to analyze the transition of the denoising level.
The visualization enables users to scrutinize the recovery region, which the model predicts based on the trajectory of inputs and outputs at each step of denoising.
Moreover, the approach of probing external models through randomized input sampling exhibits robustness concerning intrinsic issues of the model, such as gradient vanishing.
However, its applicability to generative models has limitations primarily due to two key reasons.
First, the absence of a designated target and similarity function for deriving the saliency score, such as a class or bounding box, renders such methods inapplicable. 
Second, the existing masking techniques do not encompass entire pixels but instead focus on specific objects.
Inspired by RISE, we propose DF-RISE to figure out those limitations.
Given the image representation in $t$ step $R_t$, we generate a saliency map from the U-Net model $f$ with $N$ iteration.
Initially, we generate $n$-th random masks $\mathbf{M_n}$ with our masking method to perturb the target representation of $\mathbf{R_t}$.
Subsequently, the U-Net model predicts the latent representation using both the perturbed representation $\mathbf{R_t}\odot\mathbf{M_n}$ and the original representation $\mathbf{R_t}$.
This yields two outputs, denoted as $f(\mathbf{R_t}\odot\mathbf{M_n})$ and $f(\mathbf{R_t})$, respectively.
The confidence score is then determined by assessing the similarity between the perturbed output and the original output and then element-wise multiplying with the mask $\mathbf{M_n}$.
To obtain the final saliency map $\mathbf{S}$, we repeat the aforementioned process $N$ times, incorporating normalization and high-resolution techniques.
Our tool enables visualization of the generative diffusion model through the combined application of masking and similarity functions, resulting in insightful saliency maps.
\begin{center}
\begin{algorithm}
	\caption{The framework algorithm of the DF-RISE} 
	\begin{algorithmic}[1]
        \Require{Image representation $\mathbf{R_t}$, Diffusion model $f$}
        \State $\mathbf{S}_{I,f} \leftarrow \mathbf{0}$
        \For {$interation=1,2,\ldots,N$}
            \State Generate random gaussian mask $\mathbf{M_n}$
            \State Perturb the input representation $\mathbf{R_t}\odot\mathbf{M_n}$
            \State Obtain the target output $f(\mathbf{R_t})$
            \State Acquire perturbed output $f(\mathbf{R_t}\odot\mathbf{M_n})$
            \State Obtain $weighted\ mask$ $ \mathbf{M_n} \odot s( f(\mathbf{R_t}\odot\mathbf{M_n}), f(\mathbf{R_t}) )$
            \State $\mathbf{S}_{I,f}(\lambda) \leftarrow \mathbf{S}_{I,f}(\lambda) + weighted\ mask$
        \State Normalize $\mathbf{S}_{I,f}(\lambda)$
		\EndFor\\
            \Return{$\mathbf{S}_{I,f}(\lambda)$}
	\end{algorithmic}
    \label{alg1}
\end{algorithm}
\end{center}
First, the existing masking methods aim to generate a saliency map of a specific grid of regions to address classification or object detection tasks.
In contrast to the works, all regions of the representation should be masked to visualize the attribution of all elements for the generated output.
We generate masks by sampling $N$ random binary masks with a Gaussian distribution at each pixel element $\lambda$ of size $h \times w$ and setting pixel elements below a certain threshold to zero and elements above it to one.
The perturbed representation is computed by element-wise multiplication as $f(\mathbf{R_t}\odot\mathbf{M_n})$.
Compared to the existing method, our approach generates various types of masks representing the entire latent representation so that we can obtain the confidence score of the entire pixel elements.

Second, we leverage the structural function from the Structural Similarity Index (SSIM)~\cite{wang2004image,zhao2016loss} to acquire a similarity score between the target representation $f(\mathbf{R_t})$ and perturbed representations $f(\mathbf{R_t}\odot\mathbf{M_n})$.
Therefore, we compare three concepts that are sensitive to the Human Visual System (HVS): luminance, contrast, and structure as follows:
\begin{equation}\label{eq5}
l(a,b)= \frac{2\mu_a\mu_b + C_1}{\mu_a^2 + \mu_b^2+C_1},
\end{equation}

\begin{equation}\label{eq6}
cs(a,b) = \frac{2\sigma_{ab} + C_2}{\sigma_a^2 + \sigma_b^2 + C_2 },
\end{equation}
where $C_n$ is a constant that prevents each denominator from becoming zero and $a,b$ are input images to be compared.
Luminance is a value that indicates the brightness of light, which is not suitable for predicting noise.
Since contrast also represents the contrast of color variance, this mechanism only attends to the gap between the object and the background.
Although luminance and contrast have characteristics that vary depending on the scene, the structure is independent of them and does not change depending on different views or colors:
\begin{equation}\label{eq7} 
s(a,b) = \frac{\sigma_{ab}+C_3}{\sigma_a\sigma_b + C_3}.
\end{equation}

Therefore, the similarity between noisy images should be based on the unchanged quality of the structure in each image representation.
We implement the structure similarity function to compute the similarity score about noise structure as follows:
\begin{equation}\label{eq8}
\mathbf{S}_{I,f}(\lambda) = \sum_N s( f(\mathbf{I}\odot\mathbf{M_n}), f(\mathbf{I})),
\end{equation}
\begin{equation}\label{eq9}
\mathbf{S} = \frac{\mathbf{S}_{I,f}(\lambda) - min(\mathbf{S}_{I,f}(\lambda))}{max(\mathbf{S}_{I,f}(\lambda)) - min(\mathbf{S}_{I,f}(\lambda))}.
\end{equation}

We compute the similarity score between the latent and masked latent representations of a particular step $t$ that we intend to visualize.
The saliency map $\mathbf{S}$ is computed from the normalized confidence score by min-max normalization. Without an additional backpropagation process or conditional representation, we interpret the denoising process as agnostic to the conditional representation.
Fig.~\ref{fig:2} illustrates the framework of DF-RISE. To perturb the image representation at $t$, $R_t$, $n$ number of mask $M_n$ is generated. The U-Net model $f$ predicts $f(R_t)$ and $f(R_t\odot M_n)$ from perturbed input $R_t\odot M_n$ and original input $R_t$. The similarity score between $f(R_t)$ and $f(R_t\odot M_n)$ is multiplied to make saliency map $S$.
Algorithm~\ref{alg1} outlines the aforementioned DF-RISE procedure utilized for generating the saliency map from the noise prediction model.

\noindent\textbf{DF-CAM: Diffusion gradient-weighted Class Activation Mapping.}
For a comprehensive analysis of the visual concept that the model concentrates on within an image, users necessitate a visualization of the internal workings of the model.
Grad-CAM, a visual explanation method that implements a class-discriminative localization technique for CNN-based neural networks without requiring retraining, presents explanations aligned with the model's internal workings.
Furthermore, by leveraging the gradient flow within the model, the gradient-weighted activation mapping method allows us to explain the decision process of each layer. 
However, the method lacks applicability to generative models since the generative model has no class or boundary box for the target.
To overcome this limitation, we propose DF-CAM, specifically designed for U-Net, the noise prediction network, to address issues.
We first compute the gradients from all pixels of the target feature map by using the sum of the pixels of the output representation $R_t$, as the target score.
The gradients propagating backward are globally average-pooled to calculate neuron weights $\alpha_k^{R_t}$: 
\begin{equation}\label{eq10}
\alpha_k^{R_t}=\frac{1}{\lambda}\sum_i\sum_j\frac{\partial \sum_\lambda{R_t}}{\partial A_{ij}^k},
\end{equation}
where $A^k_{ij}$ represents the value at a specific spatial location $(i, j)$ within the feature map associated with channel or neuron $k$.
In the final step, we calculate a linear combination of the neuron weights with the forward activation map, followed by the activation of a ReLU layer to extract positive influences and create a refined heatmap, denoted as $L_{DF-CAM}$:
\begin{equation}\label{eq11}
L_{DF-CAM}=ReLU(\sum_k{\alpha_k^{R_t} A^k}).
\end{equation}

Obviously, the resulting refined heatmap requires normalization and interpolation before being visualized as an image.
By implementing DF-CAM, we visualize the specific $t$ step of the denoising process using a heatmap that aligns with the neural network's architecture. 
Moreover, this enables us to perform layer-wise visualization, allowing us to analyze the distinct decision processes taking place at each layer of the network.
The heatmap provides valuable insights into the model's internal operations and aids in understanding the denoising process at different stages of the network's execution.

Our visualization methods explain the denoising process from different approaches. 
Therefore, we perform several experiments and evaluations in the ~\ref{experiments} sections to compare the faithfulness and applicability of the two tools, as well as their strengths and weaknesses.
For a baseline evaluation of our visualization method, we compare the LIME implementation for the generative model to ours with the AUC score.

\subsection{Exponential time-step sampling}\label{method2}
In contrast to the previous research issue of the visualization of the decision-making process of each inference step, we have an issue comprehending the visual concept in each stage of in the total time-step of $T$ beyond the observation of SNR.
Therefore, we analyze the last research question: What visual concept is implied at time-step $t$?
Users cannot identify which visual concept is intensive in each stage in time-step $T$, just understanding the differential SNR~\cite{choi2022perception,kwon2022diffusion}.
Instead, we implement an Exponential time-steps sampling tool that attends to a specific stage to contrast the different outputs of the diffusion model.
In the diffusion process, time steps are selected at equal intervals, resulting in a uniform distribution at each stage.
Our goal is to analyze the visual concept emphasized at each stage by sampling time steps with a higher distribution in specific intervals.
Nonetheless, restricting sampling to a particular stage could disrupt the generation process, since the diffusion model is trained across all time steps.
Therefore, the sampling method employs an exponential function to intensively sample a specific stage leaving a few sampling for other stages.
Moreover, we control the intensive stage in $T$ number of steps by the parallel transition of the exponential function as follows:
\begin{equation}\label{eq12}
\delta^{l+\gamma}=T,
\end{equation}
where $\delta$ and $l$ denote the exponent and the number of inference steps, respectively, and  $\gamma$ indicates a parallel transition variable for the exponential function. We deploy the expression to calculate $\delta$:
\begin{equation}\label{eq13}
\delta = e^{\log T/(l+\gamma)},
\end{equation}
\begin{equation}\label{eq14}
p_t = T - \delta^{t+\gamma}.
\end{equation}
Without the parallel transition ($\gamma$ is zero), $p_t$ is the $t$-th step in the sampled stage, which has a shorter interval in the specific time stage, whereas $\delta^{t}$ is the flipped sampled time-step focusing on the latter stage.
As the reverse process of diffusion starts from $T$, we subtract $\delta^{t+\gamma}$ from $T$ for the early sampling, as shown in Eq.~\ref{eq14}. This adjustment allows us to achieve finer intervals in the early stages in the range from $T$ to $0$.
By tuning the parallel transition variable, it is possible to adjust a particular stage with the shortest interval.
For instance, if the parameter $\gamma$ is less than $30$, the scheduler samples the time-steps concentrated between $1000$ and $800$. By introducing such parameter changes, we enable a comparison of the visual concepts generated at each stage through the generated images.
The diffusion model proceeds the denoising process at the time-steps sampled by exponential sampling.
As shown in Fig.~\ref{fig:6}, we contrast three representative cases of images that are generated from different stages: uniform sampling interval, early-stage sampling, and latter-stage sampling.
Specifically, in an existing reverse process within the diffusion model, they sampled a subset of $l$ time steps from a total of $T$ time steps using equal intervals of $\frac{T}{l}$, aforementioned uniform sampling (When $T$ equals $1000$, 50 intervals occurs from 20 inference steps $l$).
By the way, we set the parameter $\gamma$ to $60$ in Eq.~\ref{eq14} to get the time steps $p_t$ for the early sampling.
Latter sampling reverses the early sampling by subtracting the $T$ and computing the absolute value from Eq.~\ref{eq14}.
Comparing the outputs generated by the early and the latter stages, we interpret the visual concepts of each stage with similar attributes as illustrated in Fig.~\ref{fig:6}, such as ``a form of a cat on the football lawn", but clearly different attributes and different motions of ``playing soccer".
A more specific interpretation of visual concepts is described in Sec.~\ref{exp2}.
\label{methods}

\section{Experiments}
We propose and address three key research questions pertaining to the interpretation and analysis of the diffusion generation process, employing three distinct visualization analysis tools. 
DF-RISE highlights the denoising level in each inference step through saliency maps from external of the model, DF-CAM interprets the visual concepts the model attends to in the denoising step by visualizing the inner working, and exponential sampling provides insight into the visual concept by saturating the sampling interval to a specific time stage.
We focus on interpreting the pre-trained latent diffusion model~\cite{rombach2022high}, one of the generative diffusion models, without additional training through the model inference with $30$ steps. 
The pre-trained model was trained on $512\times512$ images from a subset of the LAION-5B~\cite{schuhmann2022laionb} database, known for its prominence as an open-source framework supported by a significant community. To ensure stable inference, we incorporated deterministic DDIM sampling as the reverse process. Additionally, the text-to-image model employed a fixed, pre-trained CLIP ViT-L/14 for the text encoder. 
During the inference process, we visualize the UNet model, which predicts noise within the latent space.
Furthermore, the Exponential sampling method introduces variations in the sampling approach that can be applied to the DDIM when sampling time step.
In Sec.~\ref{exp1}, we rigorously assess the tools to prove faithfulness with several qualitative and quantitative evaluations.
In Sec.~\ref{exp2}, We utilize our tools to interpret and analyze the decision-making process to solve the research questions aforementioned.
\begin{figure}
\begin{center}
\includegraphics[width=1.0\linewidth]{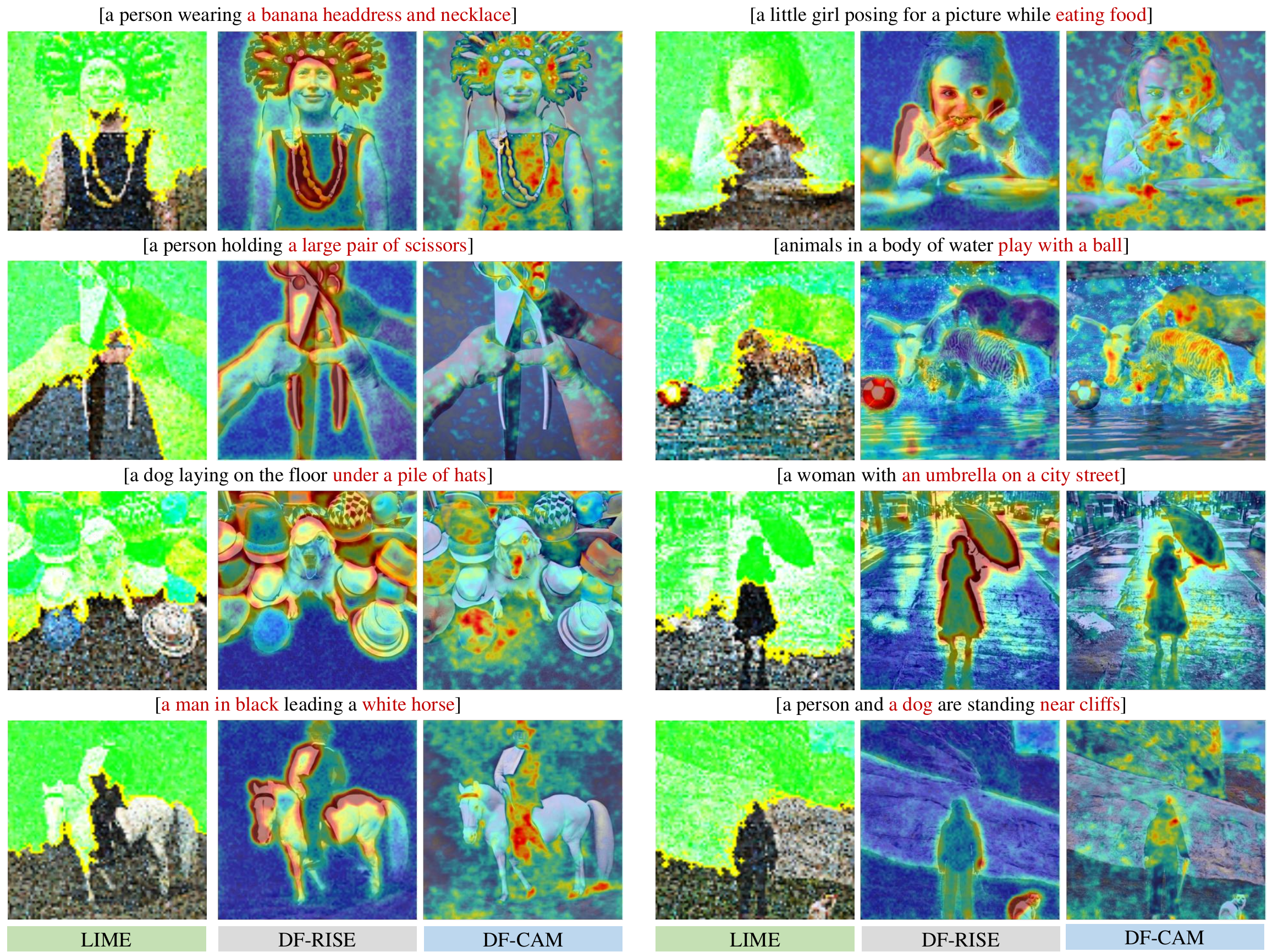}
\end{center}
\caption{\textbf{Qualitative evaluation DF-RISE with baseline} We compare DF-RISE visualization to the LIME for qualitative evaluation. While the LIME depends on the segmentation region, DF-RISE visualizes the decision process unaffected by external factors.}
\label{fig:3}
\end{figure}
\begin{figure}
\begin{center}
\includegraphics[width=0.8\linewidth]{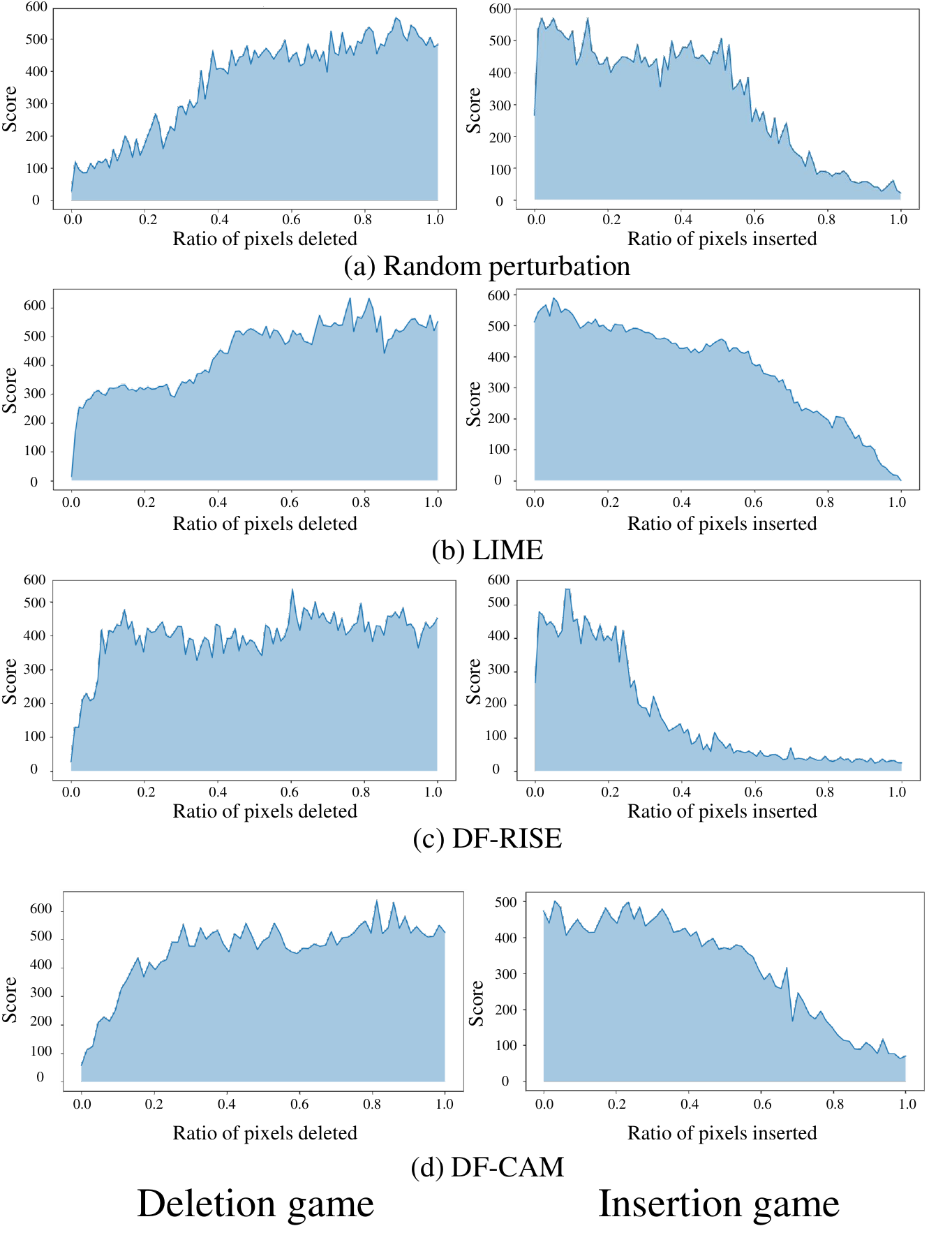}
\end{center}
\caption{\textbf{Comparisons for deletion game and insertion game with baselines}
We evaluate the deletion and insertion game for DF-CAM and DF-RISE with baseline and random perturbations. We delete or insert data from high to low activation. The deletion graph is in the first column, and the insertion graph is in the second column. The initial derivative of the curve indicates whether the key information is identified.
}
\label{fig:4}
\end{figure}
\subsection{Quantitative and qualitative evaluations.}\label{exp1}
\noindent\textbf{Saliency map visualization methods.}\label{exp1-1}
We interpret the diffusion process using DF-RISE and DF-CAM and describe the diffusion process from the perspective of time-steps. To demonstrate the faithfulness of the saliency map visualizations, we apply LIME to the attribute visualization tool for the diffusion noise prediction model as a baseline. To compare the performance with the baseline, we perform both quantitative and qualitative evaluations. For qualitative evaluation, we compare the activated areas of the model when visualizing the image generation process step by step. As shown in Fig.~\ref{fig:3}, each visualization method focuses differently on the attributes and activities. As shown in Fig.~\ref{fig:3}, LIME visualization is excessively affected by segmentation performance since the saliency map exists in the segmented region. Moreover, LIME only visualizes the activated regions within the segmentation area but does not show the relative differences in activation levels. DF-RISE and DF-CAM are not affected by other performances, such as segmentation, and express the degree of visual information that the model attends to using the heatmap. Illustrated in Fig.~\ref{fig:3}, the semantic visual concepts (red words in the prompt) are clearly activated in our method.
Specifically, the DF-CAM and DF-RISE visualize the level of denoising in each step and the feature related to the visual concept, respectively.
%
\begin{table}[]
\begin{center}
\caption{\textbf{Comparisons for AUC score with baselines.}
The AUC scores regarding the deletion (lower is better) and insertion (higher is better) games on the Karpathy test split dataset. We compare DF-CAM, DF-RISE, LIME, and random perturbation for deletion and insertion games.}
\begin{tabular}{lll}
\hline
        & Deletion ($\uparrow$) & Insertion ($\downarrow$) \\ \hline
Random  & 0.6553   &  0.6411   \\
LIME & 0.5947   & 0.6925    \\
DF-RISE & \textbf{0.8042}   & \underline{0.5933}    \\
DF-CAM  & \underline{0.7012}  & \textbf{0.5532}       \\ \hline
\end{tabular}
\label{table1}
\end{center}
\end{table}
For quantitative evaluation, we compare the AUC score~\cite{petsiuk2018rise,petsiuk2021black} with the baseline. The AUC score was calculated in two versions. We implement the FID score to compute the image similarity while ensuring the objectivity of the similarity function. First, we get the score (the lower the score, the higher the similarity) between the perturbed image and the original image by deleting the image's important attributes in order and calculating the AUC score. Second, we evaluate the FID score by sequentially inserting the important attributes of the image and calculating the AUC score. We use the $1 K$ captions of the Karpathy test split~\cite{karpathy2015deep} dataset for the prompts.

The deletion game involves systematically removing pixels based on each visualization method perceived importance, typically starting with those crucial for visual information.
Then, AUC score assesses how dropping these pixels affects the FID score.
Conversely, in the insertion game, pixels are added back into the image based on their ranked importance, revealing their impact on image generation or modification.
Both methods help evaluate the significance of pixels in generating or altering images.
As shown in Fig.~\ref{fig:4}, we validate the results of the baseline and random perturbations with DF-RISE and DF-CAM. We show that DF-CAM and DF-RISE exhibit the best performance from the steepest gradient changes in the early perturbation step in the deletion game and insertion game, respectively as shown in Fig.~\ref{fig:4} and Tab.~\ref{table1}.

\begin{table}[]
\begin{center}
\caption{\textbf{AUC score comparison for ablation studies.} We evaluate the other similarity function for image similarity score (\eg FID, cosine similarity, and contrast score) and masking method from RISE. Bold scores perform best, followed by underlined scores}
\begin{tabular}{llll}
\hline
             &       &  Deletion ($\uparrow$) & Insertion ($\downarrow$)     \\ \hline
DF-RISE         &        & \textbf{0.8042}    & \underline{0.5933} \\ \hline
similarity   & FID      & 0.6247          & 0.6250          \\
function                 & cos      & 0.6166 & \textbf{0.5850}          \\
                   & contrast & 0.6051          & 0.7563          \\ \hline
masking       & RISE     & \underline{0.6553}         &  0.6397          \\ \hline
\end{tabular}
\label{table2}
\end{center}
\end{table}
\begin{figure}
\begin{center}
   \includegraphics[width=0.8\linewidth]{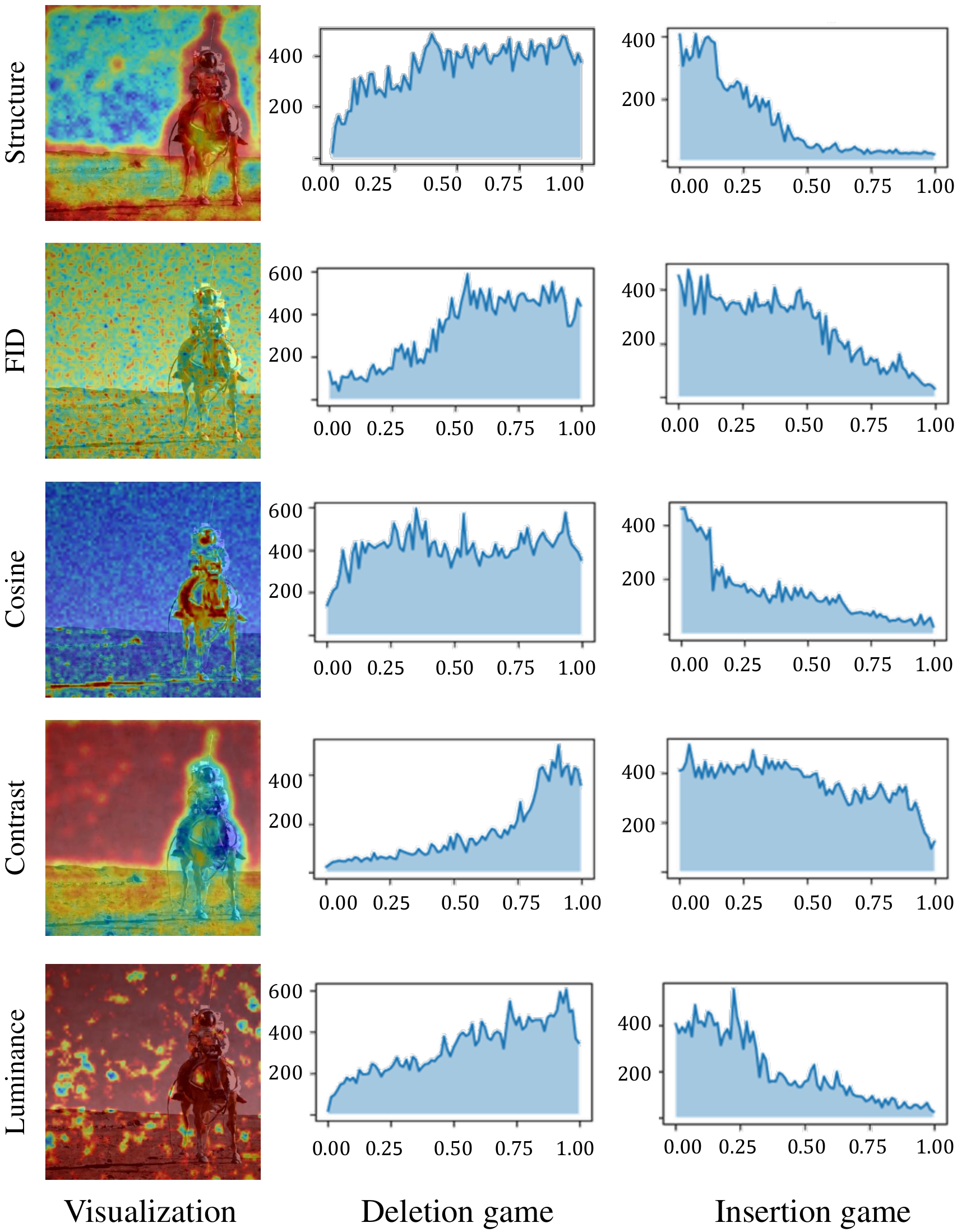}
\end{center}
   \caption{\textbf{Ablation study. }
   We generate an image of `an astronaut riding a horse on Mars'. We compare the similarity function with FID score, Cosine similarity, luminance, and contrast. The first column is the heat map output, the second column is the result of the deletion game, and the third column is the result of the insertion game.
 }
\label{fig:5}
\end{figure}
\noindent\textbf{Ablation studies.}\label{exp1-2}
To demonstrate the performance of the masking approach and the similarity function used in DF-RISE, we conducted ablation studies. First, we evaluate and compare the results of using different similarity functions to demonstrate the performance of the structural similarity function designed for the generative model. We conducted an ablation study by comparing the FID score, cosine similarity, and contrast score from SSIM. As summarized in Tab.~\ref{table2}, the cosine similarity function yields the best results in the similarity function about the insertion game. On the other hand, DF-RISE outperforms with a high gap in the deletion game. This happens since the AUC score is less sensitively affected by the low ratio of pixels inserted (\eg 0.01 point).
Second, to validate the masking method designed for the generative model saliency map, we compare it with the masking method used in RISE.
Furthermore, we evaluate our similarity function by comparing the heat map of images from various similarity score functions.
As shown in Fig.5, the result of the FID score cannot get a saliency map since it is affected sensitively by data distribution.
Although the result of the cosine function obtains adequate heatmap, the visual concept of `mars' is excepted.
Conversely, the result of structural function activates all the elements.
The contrast and the luminance do not focus on the semantic feature but on the difference in the color.
All those results show that our masking method is more applicable for generative diffusion model visualization.

\begin{figure*}
\begin{center}
\includegraphics[width=1.0\linewidth]{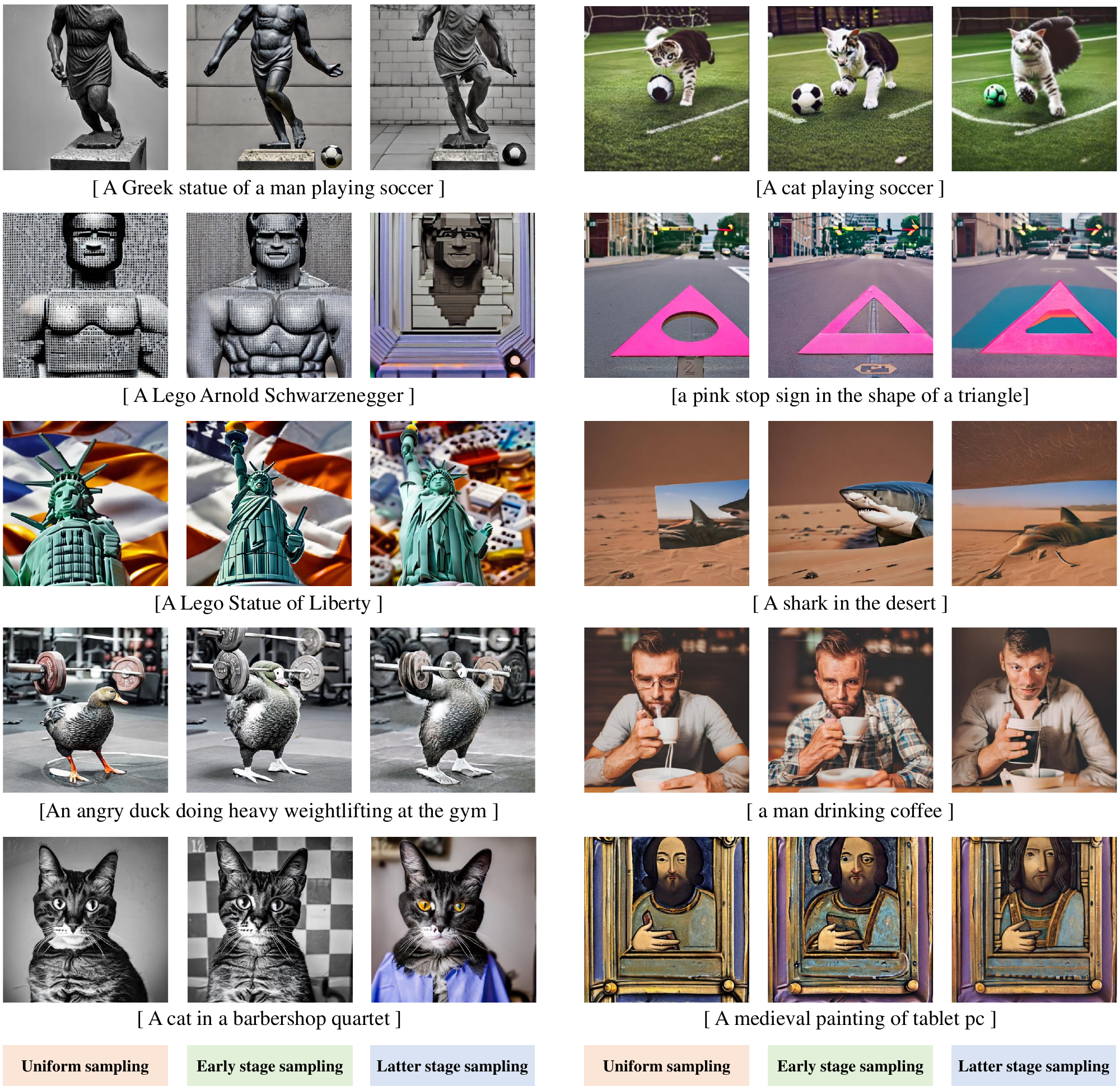}
\end{center}
   \caption{\textbf{Images generated by Exponential time-steps sampling.}
   The different images are created simply by controlling the time-steps sampling in the same $30$ inference steps. The uniform sampling samples the time-steps for a uniform interval. To compare to the uniform sampling, we implemented the exponential sampling to adjust the $\gamma$ to 60. Early sampling focuses on the early stage sampling (approximately $999$ to $700$), and latter stage sampling concentrates on the latter stage (approximately $200$ to $0$).}
\label{fig:6}
\end{figure*}
\begin{table}[]
\begin{center}
\caption{The quantification of concepts for exponential sampling method using visual concepts of subject, relation, and object from Visual Genome dataset.}
\begin{tabular}{llll}
\hline
                 & subject        & relation      & object         \\ \hline
uniform sampling & \textbf{21.61} & 2.99          & 17.61   \\
latter sampling  & \underline{21.30}     & \underline{2.40}    & \underline{17.98}          \\
early sampling   & 19.98          & \textbf{3.04} & \textbf{18.51} \\ \hline
\end{tabular}
\label{table3}
\end{center}
\label{table4}
\end{table}

\noindent\textbf{Exponential time-step sampling.}\label{exp1-3}
Although the exponential time-step sampling visualizes the divergence of visual concepts through image outputs as illustrated in Fig.~\ref{fig:6}, the tool does not still assure faithfulness.
To address this, we introduce a quantification of the concept that addresses the correlation between visual concepts and the time stage.
To facilitate the evaluation, we leveraged the Visual Genome dataset~\cite{krishna2017visual} to construct a prompt test dataset.
We generate $50$ prompt sentences, and each sentence is composed of subject, relation, and object words from the Visual Genome dataset.
For instance, the word 'car' is a visual concept of the subject, 'moving down' is the relation, and 'the street' is a visual concept of the object from `car moving down the street'.
Moreover, we quantify the visual concept score by employing the relevance score~\cite{chefer2021generic} between the text representation from the text encoder and the image representation from the U-Net model.
As illustrated in Tab.~\ref{table3}, the relevance score between the image and the visual concept composed of subject, relation, and object indicates the distinctive visual concept trends of each time stage.
Compared to uniform sampling which emphasizes the visual concept of the 'subject', an early stage has a higher level of the visual concept of 'relation' or 'object'.
Therefore, we prove that the visual concept level that differs from the sampling stage could be computed by quantification of the visual concept.
The quantification validates the likelihood of measuring the distinct tendency of visual concepts across time-steps using exponential sampling.
\subsection{Visual analysis of the diffusion generation process}\label{exp2}
\begin{figure*}
\begin{center}
\includegraphics[width=1.0\linewidth]{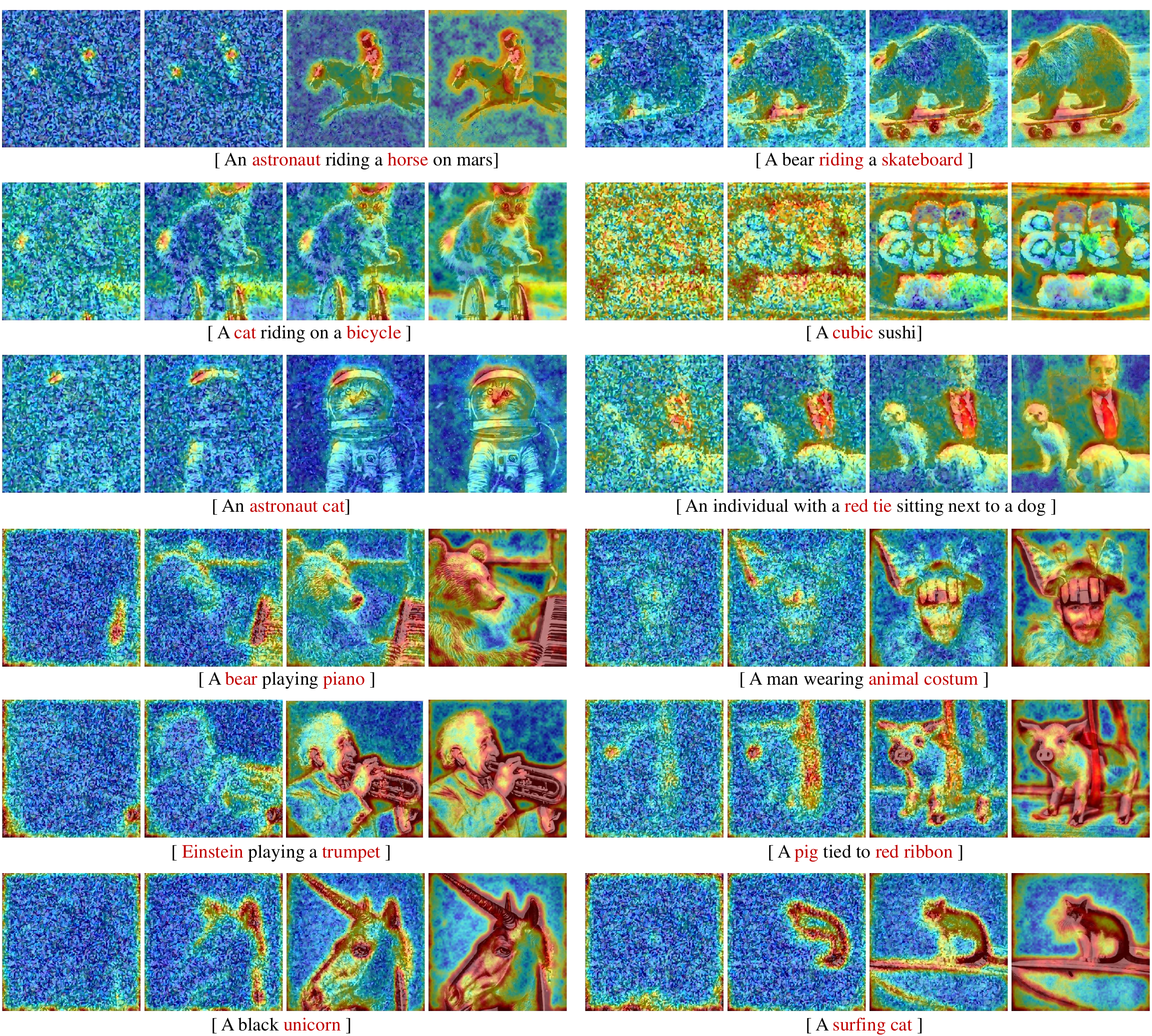}
\end{center}
\caption{\textbf{Visualization of the denoising level at each inference process employing DF-RISE.}
We visualize all steps of the denoising process using DF-RISE (from the left side to the right side). To observe notable changes, we illustrated four representative DF-RISE out of 30 inference steps in the figure.} In the early stage, the model concentrates on the structure of visual concepts. By performing the latter step, we confirm that the area on which the model activates becomes diffused.
\label{fig:7}
\end{figure*}
\noindent\textbf{The denoising region is diffused from the structure to the fine-grained area, visualized DF-RISE.}\label{exp2-1}
We visualize the region in which the diffusion model activates for each step in the noisy image. We generate images with the same time-step sampling interval and visualize the area where the diffusion model concentrates at each step. As illustrated in Fig.~\ref{fig:7}, in the early stages, we confirm that the head, the most meaningful part of the words `horse' and `astronaut' in the prompt, is activated first. Subsequently, the activated area gradually diffuses, starting with the astronaut's body. Various other results demonstrate that the diffusion process starts from semantic structure regions (`horse's head', `bicycle's handle', `bear's head', and `cubic shape') and diffuses to each detail region of elements (`mars', `riding', `skateboard', and `cubic sushi'). Through these experimental results, we visually explain and answer the question about the denoising process of the diffusion process:
What regions are recovered by the diffusion model in terms of semantic and detail levels?
Compared with the drawing process by humans, which starts from the border to the details of elements, diffusion models start from a semantic structure in the conditional prompt.
In the early process, the diffusion model creates a semantic structure and then widens to the details of the elements.
From the visual analysis of the saliency map of the denoising level, we observe the part of the structure that the model tends to generate and the denoising process in each step.
Therefore, the semantic level of the denoising process is the spatial information that represents the structure of the objects and the detail level denotes a fine-grained region in all pixels.
\begin{figure*}
\begin{center}
\includegraphics[width=0.9\linewidth]{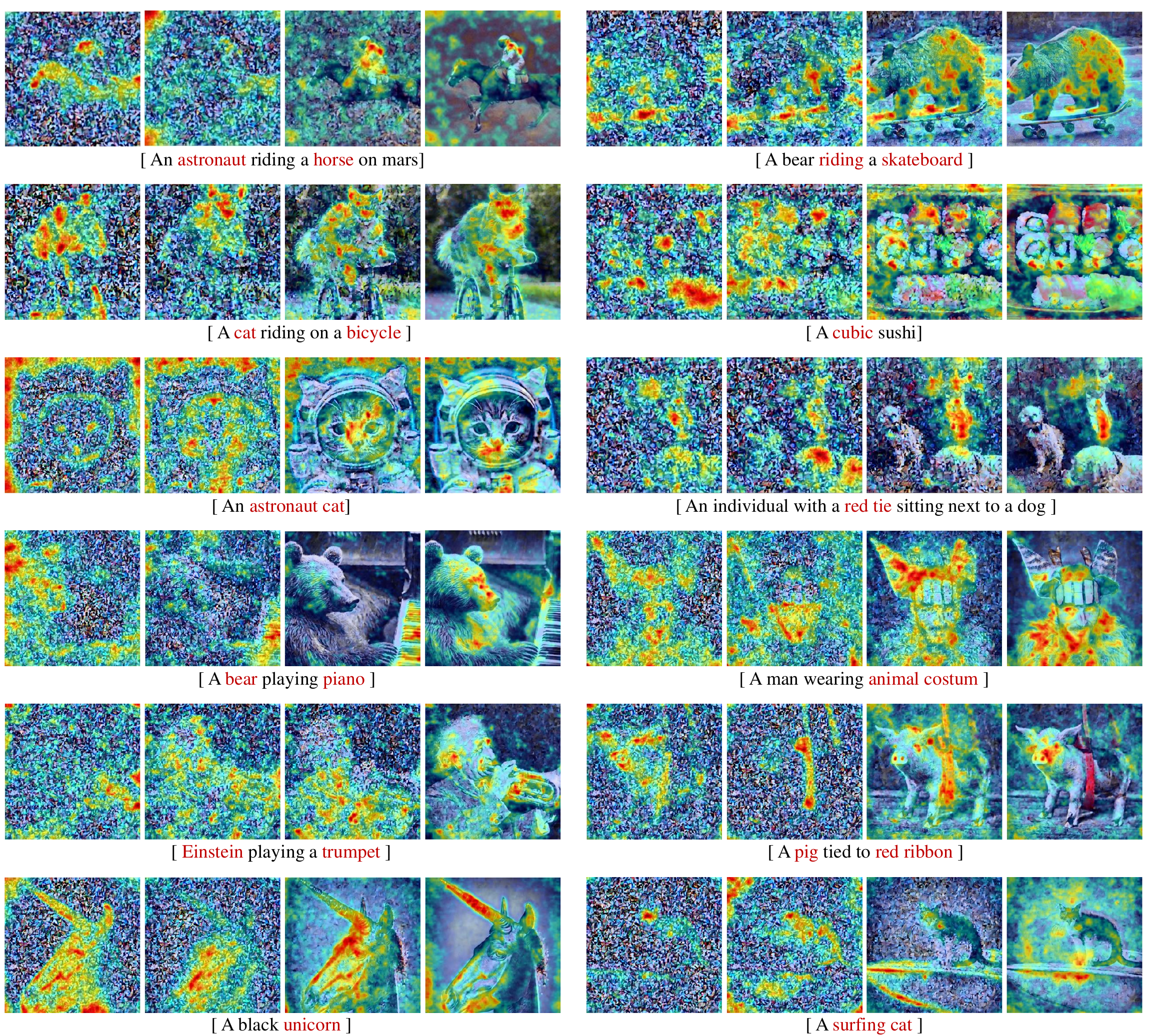}
\end{center}
\caption{\textbf{Visualization of the feature model predicts at each denoising process employing DF-CAM.}
We visualize some steps of the denoising process using DF-CAM (from the left side to the right side). The diffusion model concentrates on different concepts in each step to generate optimal output. The saliency map helps users to understand the decision-making process to realize the concepts in the prompt.}
\label{fig:8}
\end{figure*}

\noindent\textbf{The diffusion model focuses on different visual concepts in each denoising step, visualized by DF-CAM.}\label{exp2-2}
In the previous work, we visualize the different levels of denoising utilizing the DF-RISE and interpret the noise change.s
In contrast to DF-RISE, which visualizes the external of the model, we analyze the visual concept that the denoising model attends to by visualizing the internal workings of the model with DF-CAM.
As illustrated in both Fig.~\ref{fig:7} and Fig.~\ref{fig:8} which represent identical samples with distinct visualization, DF-RISE exhibits the denoising level diffuses with each subsequent step, and DF-CAM visualizes the inner workings of the model concerning the visual concept.
The activated region indicates the visual concept in DF-CAM, which is a result of the decision-making process of the denoising model to generate the final image.
Through the analysis, we demonstrate the faithfulness of the saliency map and answer the research question $2$: Which specific concepts are prioritized at each time-step in order to generate images that align with the given conditional prompt?
The denoising model predicts different visual concepts at each stage such as `astronaut', `riding', and `horse' from `an astronaut riding a horse'.
Moreover, from the step visualization, the model not only predicts the object but also focuses on the non-object concept such as background since the generation process needs a detailed denoising process for high-fidelity image output.
Moreover, we interpret how the model realizes the visual concept for the image using a saliency heat map.
For instance, the head and tail of the horse are generated mainly in the early step to generate the horse, and the model concentrates on the saddle to realize the 'riding' concept.
The visualization method helps users interpret how the model materializes some of the abstract concepts such as `riding', `sitting', and `playing'.
\begin{figure*}
\begin{center}
\includegraphics[width=0.8\linewidth]{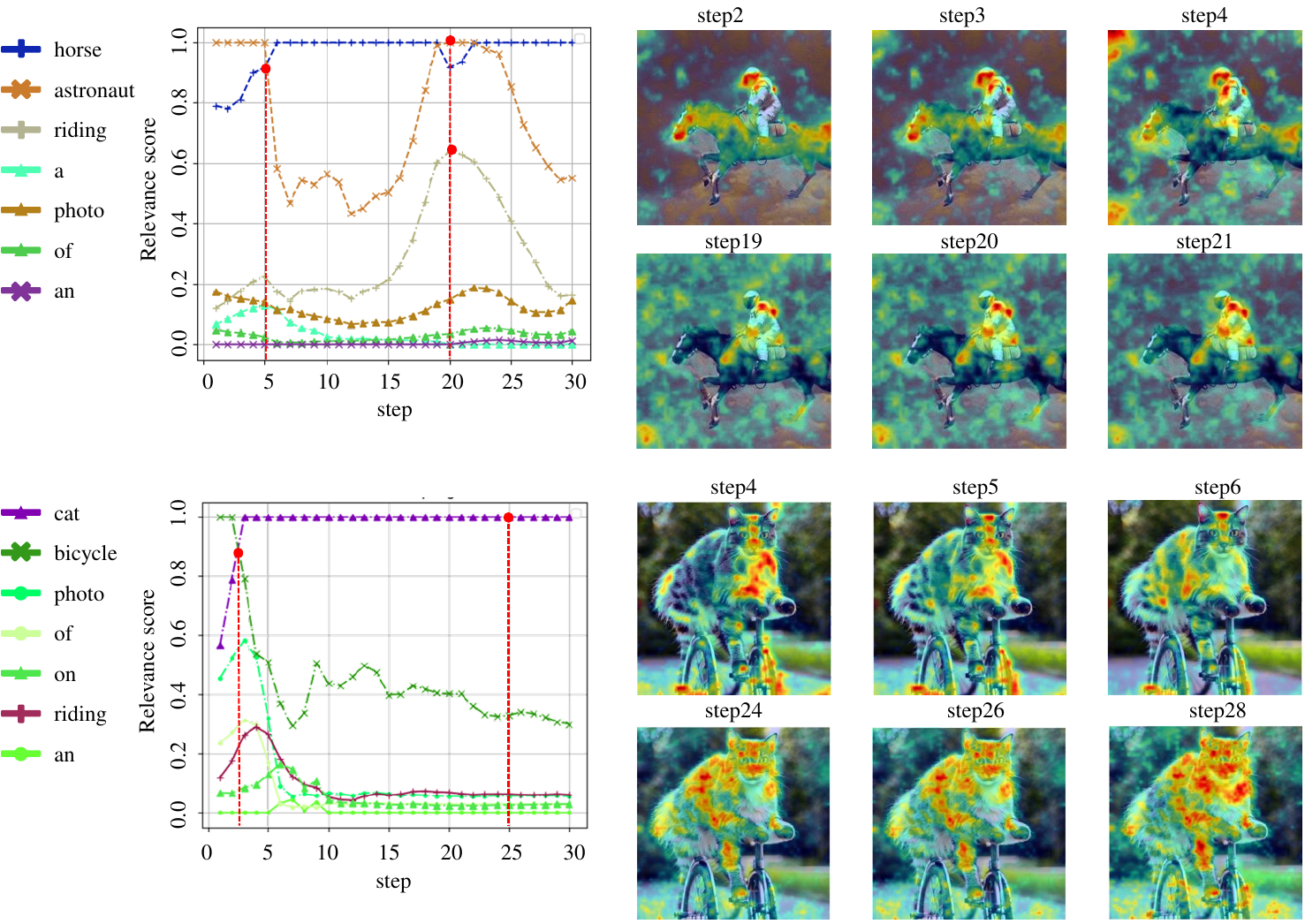}
\end{center}
\caption{\textbf{Relevance map with DF-CAM.} The left line graph indicates the relevance score between the image and prompt in each step. To prove the activated region by in the saliency map, we compare to the relevance map whether the concept exactly activated in high relevance score.}
\label{fig:9}
\end{figure*}
Moreover, we compare the relevance score with the saliency visualization to prove that the saliency map activates the visual concept the model attends to.
The line graph shows the variation in relevance scores with respect to time steps, and we select some samples of several steps that are meaningful to analyze, such as saddle or cross points.
As depicted in Fig.~\ref{fig:9}, we select steps $2$ and $4$ due to a distinct decrease and subsequent increase in the relevance score for the concepts `horse' and `astronaut', respectively.
We observe that the model shifts its focus from the `astronaut' to the `horse' as the steps progress from $2$ to $4$, by comparing the saliency map change with graph change.
As the `riding' concept increases with `horse' in the line graph, the model attends to the saddle and reins of the horse in the saliency map.
Moreover, the cross point and the highest point in the second line graph show concept change (`bicycle' and `cat') and concept attention (`cat'), respectively.
Through the analysis, we demonstrate the faithfulness of the saliency map and answer the research question $2$.

\noindent\textbf{Time-step dependent emphasis on visual concepts.}\label{exp2-3}
We visualize and interpret the denoising step from the several time-steps that are sampled by the diffusion scheduler.
Furthermore, we extend the work that focuses on only the inference steps to all time-steps in $T$.
We implement exponential sampling to concentrate on specific time-stage, and relevance scores to quantify the concept level in each stage.
First, to explain the fact that visual concepts differ across each stage, we divide the time-steps $T$ into early and latter stages and compare them.
We use pre-trained latent diffusion to generate and compare images through $30$ inference steps and the model generates images in different sampled time-steps.
The results in Fig.~\ref{fig:6} show that when sampling is mainly at a specific stage, we observe the image outputs with prominently different visual concepts preserving the main structure (\eg position or the pose of the object).
The various samples of image output provide an abundant understanding of the visual concept entailed in each time-step.
In the case of uniform sampling, the image shows the even level of feature from the prompt.
In comparison to uniform sampling, the early and latter stage of sampling produces an image that is biased toward a specific visual concept.
For instance, comparing each sampling, the attribute of 'headphone' is detected in the early stage, and the feature of gender is changed in the latter stage for the prompt `a doctor singing a song' in Fig.~\ref{fig:1}.
Likewise, the features such as `the Greek statue', `playing soccer', `Arnold Schwarzenegger', `triangle`', `Statue of Liberty', `shark', `weightlifting', `drinking', `barbershop ', and `tablet pc' are intensified, as illustrated in Fig.~\ref{fig:6}.
In contrast, the latter sampling has a more intensive level of the concepts of `soccer', `cat', `Lego', `stop sign', `A Lego', `desert', `duck', `coffee', `quartet', and `medieval'.
Through the analysis of the image concept, users interpret the visual concepts differently activated in each step just by controlling the sampling technique.
\begin{figure*}
\begin{center}
\includegraphics[width=0.9\linewidth]{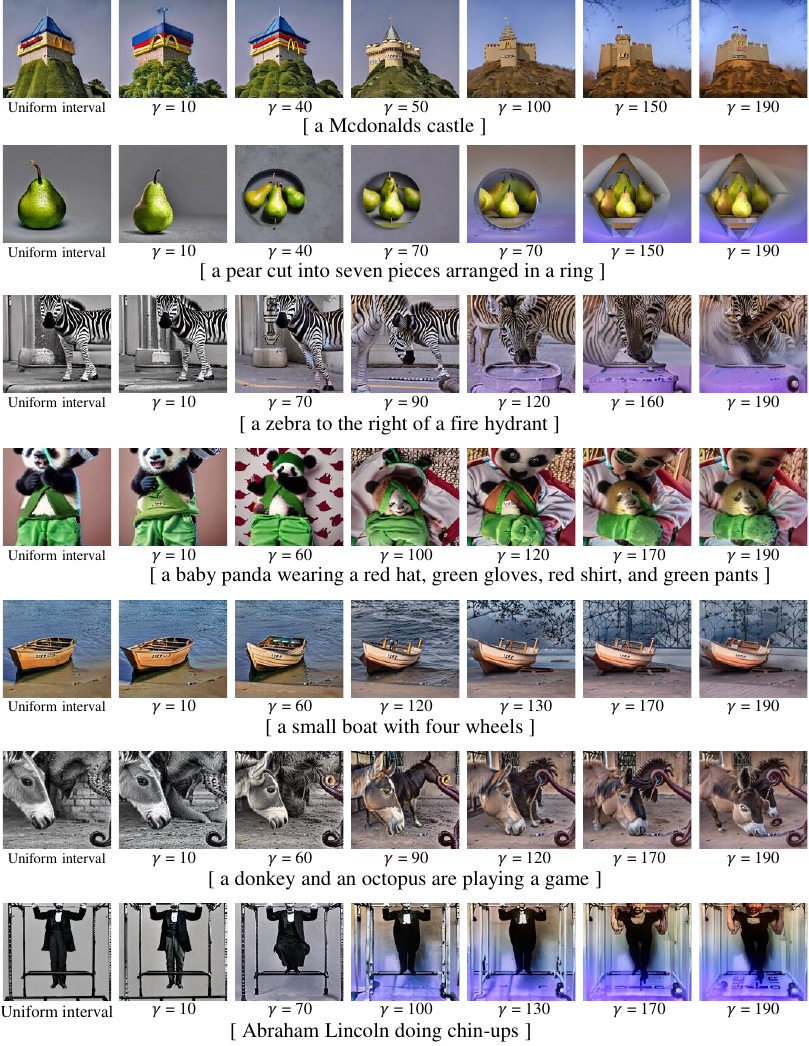}
\end{center}
\caption{\textbf{Image variation using the parallel transition of the controllable Exponential time-steps sampling.} We compare images by changing a $\gamma$ of exponential function. As users adjust the parameter, the level of visual concepts is changed by controlling the sampling stage.}
\label{fig:10}
\end{figure*}

Furthermore, we adjust the exponential sampling to analyze the difference in concepts in each stage delicately.
We demonstrate a controllable scheduler through a parallel transition (the change in $\gamma$) in Eq.~\ref{eq14}.
By varying the parameters, we adjust the stage to focus on the sampling at each time-step.
As the parameter increases, the stage moves towards $0$ at T. Fig.~\ref{fig:10} shows the image changes based on $\gamma$ variation. The visual concepts of `McDonald's', `a pear', `a zebra', `a panda', `a boat', and `Abraham Lincoln' is more activated in the early stage due to the low value of the parameter. In the latter stage, the visual concept of `castle', `seven pieces', `hydrant', `baby, red shirt', `wheels', `playing a game', and `chin-ups' are more activated, respectively.
Even with a simple sampling stage change in the same inference steps, we observe significantly different generated images.
From these analyses, we interpret the visual concepts entailed in each time-step with the following research question: what visual concept is implied at time-step $t$?
Moreover, by using the controllable scheduler, we interpret the attribute that we want to activate at each stage and discover the stage for generating optimally balanced attributes of an image, and it can be used for image editing to a specific prompt for future works.
\begin{figure*}
\begin{center}
\includegraphics[width=1.0\linewidth]{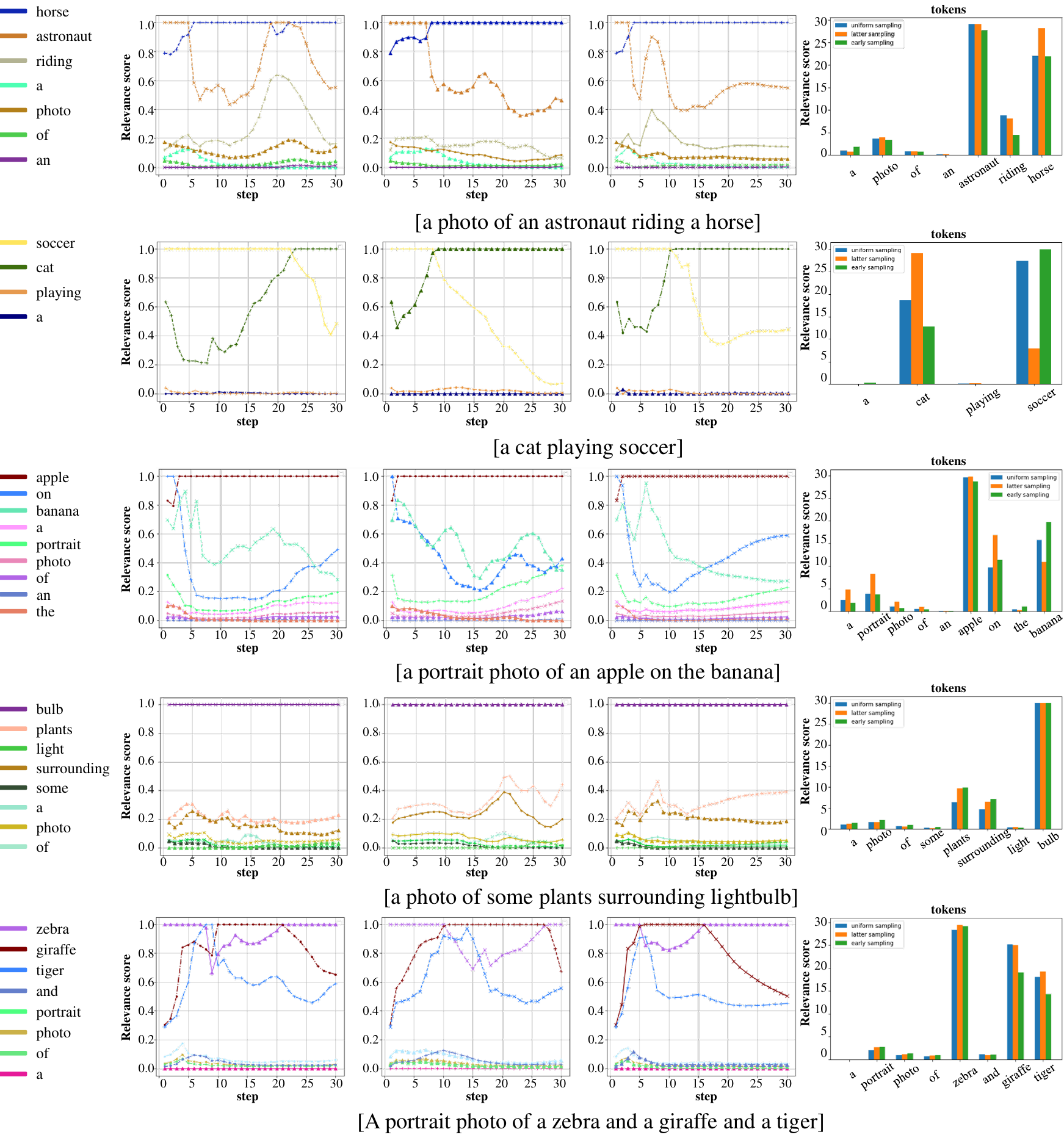}
\end{center}
\caption{Relevance map of visual concepts. The line graphs show a relevance score of each concept at time-step. The graphs include uniform sampling, early sampling, and latter sampling. A block graph illustrates the sum of the relevance score of all time-steps.}
\label{fig:11}
\end{figure*}

\noindent\textbf{Analyzing subjectivity in the interpretation of visual concepts entailed in time-steps.}\label{exp2-4}
Some of the interpretation of the image output is subjective and uncertain for users since one visual concept has various features such as `headphones' or an `opening mouse' for the concept of `singing', as shown in Fig.~\ref{fig:1}.
We implemented the relevance score~\cite{chefer2021generic} to the objectification of relevance between the visual concept in the prompt and image.
We construct several formats of prompts, including 'objects having low correlation', 'subject doing something', 'the location of two objects', 'description feature of object', and 'multiple objects' for the representative comparison.
The line graph in Fig.~\ref{fig:11} depicts the relevance score of the visual prompt about uniform sampling, early sampling, and latter sampling, respectively.
The block graph indicates the total relevance score of each visual concept across all steps.
For the analysis of the relevance score, the user can understand the visual concept level that the model concentrates on in each inference step and each time stage. 
As illustrated in Fig.~\ref{fig:11} the diffusion model mainly focuses on concepts such as objects, relations, and verbs.
Specifically, the early and latter stages in the first sample generate images by concentrating on the objects (not `riding' but `horse' or `astronaut') compared to uniform sampling.
From the block graph of the relevance score of concepts, the `horse' is emphasized in the latter sampling, and the `astronaut' has a lower score in the latter stage.
Furthermore, as depicted in the uniform sampling line graph, the `riding' and `horse' concepts have a parallel tendency, like human understanding.
The interpretation shows us the aligned visual concept that has high relevance such as riding a horse, and different relevance scores in each step.
Likewise, we categorize the concepts that each stage mostly attends to in the visual analysis.
From the overall analysis, the early stage mostly focuses on `object' in the format of `subject doing object', the latter location in the format of `the location of two objects', the background of the image in the format of `description feature of object', and fewer concentrates on the objects in the 'multiple objects' format.
Compared to the early stage, the latter stage focuses on the `subject',` location information, `object', and `multiple objects'.
We observe that these differentially emphasized visual concepts have an impact on the final image.
Based on these analyses, we extend our interpretation beyond research question $3$ that the visual concepts entailed in each step play a role as evidence to produce the respective final images.
\label{experiments}
\section{Conclusion}
In this study, we devised $3$ distinct research questions aimed at elucidating the denoising mechanism inherent in the diffusion model.
We visualized the black-box denoising process by using visualization tools including DF-RISE, DF-CAM, and exponential time-step sampling.
Our approach involves a comprehensive visual analysis that tackles the inquiries head-on.
Moreover, we performed both quantitative and qualitative evaluations to establish the faithfulness of the tools and quantified the visual concept to objectify the outcomes of sampling a relevance score.
By employing these efficacious tools, we not only shed light on the denoising progress at each discrete step but also explained the visual concepts synthesized by the diffusion model. Additionally, we interpreted the intricate concepts embedded within various time-steps, rendering them comprehensible to human cognition.
However, the robustness of our approach is challenged by the diverse forms of prompts encountered, and the saliency map still needs refinement to facilitate a more exhaustive visual analysis.
Furthermore, future research should explore the utilization of sampling methods beyond visualizing at the visual concept level, extending to editing nonsensical concepts that arise in image generation.
Overall, our research not only contributes significant insights into understanding the diffusion process but also paves the way for future investigation into the explainable diffusion process.
\label{conclusion}

\section*{Acknowledgments}
This work was supported by Institute of Information \& communications Technology Planning \& Evaluation (IITP) grant funded by the Korea government (MSIT) (No. 2019-0-00079, Artificial Intelligence Graduate School Program (Korea University) and No. 2022-0-00984, Development of Artificial Intelligence Technology for Personalized Plug-and-Play Explanation and Verification of Explanation).

\bibliographystyle{unsrt}  
\bibliography{references}

\end{document}